%% file: main.tex
\documentclass[letterpaper, 10 pt, conference]{ieeeconf}  

\usepackage{graphicx}
\usepackage{amsmath}
\usepackage{amssymb}
\usepackage{csquotes}
\usepackage{url}
\usepackage{hyperref}
\usepackage{subcaption}
\usepackage{multirow}
\usepackage{color, colortbl}
\usepackage{algorithm}
\usepackage{algpseudocode} 
\usepackage[nolist]{acronym} 
\let\labelindent\relax
\usepackage[inline]{enumitem} 
\usepackage{cite}
\graphicspath{ {imgs/} }
\usepackage{siunitx}
\newif\ifshowComments
\showCommentstrue

\ifshowComments
    \usepackage[draft,markup=bfit, deletedmarkup=sout, authormarkup=brackets]{changes}
\else
    \usepackage[final,markup=bfit, deletedmarkup=sout, authormarkup=brackets]{changes}
\fi

\definechangesauthor[name={Lukas},color=green]{LR}
\definechangesauthor[name={Matej}, color=orange]{MH}
\definechangesauthor[name={MalyMatej}, color=blue]{MM}

\ifshowComments

\else

\fi

\input{acronyms.tex}

\input{macro.tex}

\useVspacefalse 

\usepackage[firstpageonly=true]{draftwatermark}

\SetWatermarkAngle{0}
\SetWatermarkColor{black}
\SetWatermarkLightness{0.5}
\SetWatermarkFontSize{9pt}
\SetWatermarkVerCenter{30pt}
\SetWatermarkText{\parbox{30cm}{%
\centering This is the final version of the manuscript accepted to IEEE-RAS International Conference on Humanoid Robots (Humanoids), cite as:\\
\centering Rustler, L.; Misar, M. \& Hoffmann, M. (2024),\\
\centering Adaptive Electronic Skin Sensitivity for Safe Human-Robot Interaction \\
\centering  in '2024 IEEE-RAS International Conference on Humanoid Robots (Humanoids)', pp. 475-482. (C) IEEE \\
\centering \url{https://doi.org/10.1109\%2FHumanoids58906.2024.10769602}
}}

\title{Adaptive Electronic Skin Sensitivity for Safe Human-Robot Interaction}

\author{Lukas Rustler, Matej Misar, and Matej Hoffmann %
\thanks{This work was co-funded by the European Union under the project Robotics and Advanced Industrial Production (reg. no. CZ.02.01.01/00/22\_008/0004590). L.R. was additionally supported by the Grant Agency of the Czech Technical University in Prague, grant No. SGS24/096/OHK3/2T/13.}
\thanks{Lukas Rustler, Matej Misar, and Matej Hoffmann are with the Department of Cybernetics, Faculty of Electrical Engineering, Czech Technical University in Prague,
 {\tt\small lukas.rustler@fel.cvut.cz, matej.hoffmann@fel.cvut.cz.}}
 \thanks{The authors would like to thank Michael Zillich from Blue Danube Robotics for his assistance with the robotic skin.}}
 
\IEEEoverridecommandlockouts
\begin{document}

\maketitle

\begin{abstract}
Artificial electronic skins covering complete robot bodies can make physical human-robot collaboration safe and hence possible. Standards for collaborative robots (e.g., ISO/TS 15066) prescribe permissible forces and pressures during contacts with the human body. These characteristics of the collision depend on the speed of the colliding robot link but also on its effective mass. Thus, to warrant contacts complying with the Power and Force Limiting (PFL) collaborative regime but at the same time maximizing productivity, protective skin thresholds should be set individually for different parts of the robot bodies and dynamically on the run. Here we present and empirically evaluate four scenarios: (a) static and uniform – fixed thresholds for the whole skin, (b) static but different settings for robot body parts, (c) dynamically set based on every link velocity, (d) dynamically set based on effective mass of every robot link. We perform experiments in simulation and on a real 6-axis collaborative robot arm (UR10e) completely covered with sensitive skin (AIRSKIN) comprising eleven individual pads. On a mock pick-and-place scenario with transient collisions with the robot body parts and two collision reactions (stop and avoid), we demonstrate the boost in productivity in going from the most conservative setting of the skin thresholds (a) to the most adaptive setting (d). The threshold settings for every skin pad are adapted with a frequency of 25 Hz. This work can be easily extended for platforms with more degrees of freedom and larger skin coverage (humanoids) and to social human-robot interaction scenarios where contacts with the robot will be used for communication.

\end{abstract}

\input{Sections/introduction}

\input{Sections/related_work}
\input{Sections/method}
\input{Sections/experiments}
\input{Sections/conclusion}

\bibliographystyle{IEEEtran}
\bibliography{refs}

\end{document}

%% file: acronyms.tex
\newacro{dnn}[DNN]{Deep Neural Network}
\newacro{fcn}[FCN]{Fully Convolutional Network}
\newacro{pc}[PC]{Point Cloud}
\newacro{sdf}[SDF]{Signed Distance Function}
\newacro{cnn}[CNN]{Convolutional Neural Network}
\newacro{gnn}[GNN]{Graph Neural Network}
\newacro{dl}[DL]{Deep Learning}
\newacro{ml}[ML]{Machine Learning}
\newacro{gpis}[GPIS]{Gaussian Process Implicit Surface}
\newacro{mc}[MC]{Monte Carlo}
\newacro{mlp}[MLP]{Multi-Layer Perceptron}
\newacro{bpa}[BPA]{Ball Pivoting Algorithm}
\acrodefplural{gpis}[GPIS's]{Gaussian Process Implicit Surfaces}
\newacro{gpisp}[GPISP]{Gaussian Process Implicit Shape Potential}
\newacro{gp}[GP]{Gaussian Process}
\acrodefplural{gp}[GPs]{Gaussian Processes}
\newacro{ros}[ROS]{Robot Operating System}
\newacro{icp}[ICP]{Iterative Closest Point}
\newacro{fps}[FPS]{Farthest Point Sampling}
\newacro{igr}[IGR]{Implicit Geometric Regularization for Learning Shapes}
\newacro{js}[JS]{Jaccard similarity}
\newacro{cd}[CD]{Chamfer distance}
\newacro{nerf}[NeRF]{Neural Radiance Fields}
\newacro{hri}[HRI]{Human-Robot Interaction}
\newacro{phri}[pHRI]{Physical Human-Robot Interaction}
\newacro{rrmc}[RRMC]{Resolved-Rate Motion Control}
\newacro{sd}[SD]{Standard Deviation}
\newacro{pfl}[PFL]{Power and Force Limiting}
\newacro{ssm}[SSM]{Speed and Separation Monitoring}
\newacro{snn}[SNN]{Spiking Neural Network}
\newacro{arns}[aRNS]{artificial Robot Nervous System}

%% file: macro.tex
\newcommand{\equationref}[1]{\hyperref[#1]{Eq.~\ref*{#1}}}
\newcommand{\figref}[1]{\hyperref[#1]{Fig.~\ref*{#1}}}
\newcommand{\tabref}[1]{\hyperref[#1]{Table~\ref*{#1}}}
\newcommand{\secref}[1]{\hyperref[#1]{Section~\ref*{#1}}}
\newcommand{\algoref}[1]{\hyperref[#1]{Alg.~\ref*{#1}}}

\newcommand{\norm}[1]{\left\lVert#1\right\rVert}

\newcommand{\etal}[1]{#1 \textit{et al.}}

\newif\ifuseVspace
\newcommand{\customVspace}[1]{\ifuseVspace \vspace{#1} \fi}

\def\stop{\textit{STOP}}
\def\avoid{\textit{AVOID}}

\def\same{\textit{UNIFORM}}
\def\diff{\textit{BODY PARTS}}
\def\norm{\textit{LINK VELOCITY}}
\def\mass{\textit{EFFECTIVE MASS}}

\def\airskin{AIRSKIN}

%% file: Sections/introduction.tex
\section{Introduction}

\begin{figure}[t]
    \centering
    \includegraphics[width=1\columnwidth]{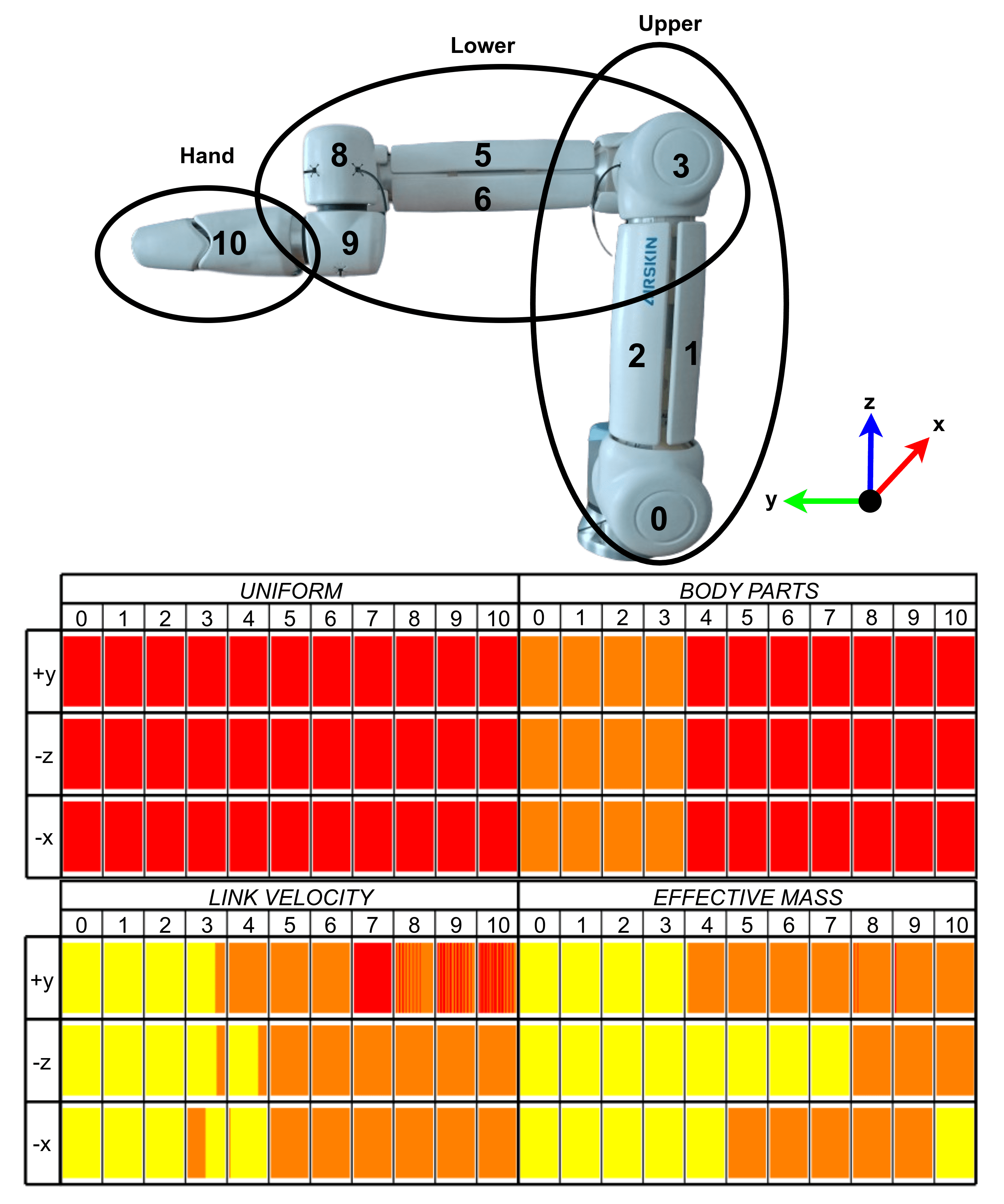}
    \caption{The UR10e robot with \airskin{} robotic skin. The ellipses distinguish three parts of the robot---upper arm, lower arm and hand. The numbers on individual skin pads represent their IDs (pads 4 and 7 on the other side of the robot). The table shows sensitivity thresholds values changing over time (with frequency of \SI{25}{\hertz}) during the motion in given axis (in robot orientation system) during a task for all the skin pads in four different settings of the robotic skin. The thresholds are color-coded based on their sensitivity as: the highest sensitivity---red, medium sensitivity---orange, lowest sensitivity---yellow.}
    \label{fig:main_fig}
    \customVspace{-1.5em}
\end{figure}


The importance of endowing robots with touch has been recognized for several decades and a large number of technologies have been developed. The focus has largely been on tactile sensing for manipulation, as equipping robot hands/fingers only with tactile sensors requires only relatively small patches of electronic skin. Whole-body artificial skins for robots have been less common, with only few successful solutions such as \cite{mittendorfer2011humanoid,mittendorfer2012integrating, park2024_LowCostSkin} or those coming from the ROBOSKIN project \cite{cannata2008embedded} and deployed most notably on the iCub humanoid robot \cite{maiolino2013flexible}. A more recent example is the Punyo-1 robot~\cite{Goncalves2022_Punyo}. The potential of sensing over the whole body surface has also not been fully explored, but \etal{Cheng} provide an overview of applications of their electronic skin  \cite{cheng2019comprehensive}. Artificial skin solutions are also starting to find their way to the collaborative robot industry through, for example, \airskin{} (detecting contact) or Bosch APAS (using proximity).

In humans, pain serves as a protective mechanism, alerting the body to potential harm and triggering a response to mitigate or avoid damage. The main receptors are nociceptors, which respond to potentially harmful stimuli and transmit this information to the nervous system~\cite{kandel1991principles}. 
The biomechanical thresholds that \enquote{trigger} pain sensations differ among individuals and are not uniform across the body. The given thresholds are usually called onsets of pain and are represented as the maximum permissible pressure before the sensation of pain appears \cite{ISOTS15066,onsetOfPain}. For example, the pressures for human arm range from 100 to $\SI{300}{\newton\per\centi\meter\squared}$.

To date, collision thresholds for artificial tactile arrays have been set uniformly. Here we present several ways in which the thresholds can be set individually on different parts of the robot bodies and dynamically on the run, ensuring safety and maximizing productivity at the same time. 
We ask ourselves the following questions: 
\textit{Should there be different sensitivity (``onsets of pain'') thresholds for different parts of the robot skin? Based on what criterion should the onsets be set?} 

This work is mainly motivated by safe \ac{hri}, namely how to set the skin sensitivity thresholds such that safety is warranted as per the Power and Force Limiting regime (PFL) of \ac{hri} \cite{ISOTS15066} but maximizing the robot performance (speed).  Specifically, we worked with the UR10e robot equipped with the \airskin{} robotic skin---see \figref{fig:main_fig}.
Collision characteristics (contact force, duration, etc.) depend on the speed of the colliding robot link but also on its
effective mass, which is a function of the whole robot configuration and the impact direction---the robot is more ``massy'' in certain configurations and collision directions.  Thus, to warrant contacts complying with the
PFL limits but at the
same time maximizing productivity, collision thresholds that will trigger robot stop or avoidance should be set individually for different parts of the robot bodies
and dynamically on the run.

The simulation, code and recorded experiment data are publicly available at \url{https://github.com/ctu-vras/adaptive-skin} and \url{https://osf.io/3adrm/}. The video is available at \url{https://youtu.be/kWt2J1Qgzpk}.


%% file: Sections/related_work.tex
\section{Related work}
This Section provides a brief overview of selected works in \ac{phri}, protective electronic skins, and ``artificial sense of pain''. 

Physical HRI, or close collaboration of humans and robots in the same physical space, is developing rapidly (e.g., ~\cite{Haddadin2016}). There are multiple modes of \ac{phri} (see, for example, \cite{Villaini2018Survey}). We are mainly interested in the so-called \ac{pfl}, described also in the safety standard of ISO/TS 15066~\cite{ISOTS15066}, where collisions between humans and robots are allowed when the force, pressure, and energy are within the prescribed limits. However, the limits are widely debated, as empirical tests show that they may be too strict in the real-world scenarios~\cite{Fischer2023CollisionTests}, or the estimated mass used in the calculation is imprecise\cite{kirschner2021effmass}.

The \ac{pfl} regime can be also combined with other \ac{phri} modes, such as \ac{ssm}, further boosting productivity~\cite{lucci2020combining, svarny2022functional}.
A different approach lies in using additional sensors. \etal{Xu}~\cite{Xu2023ActiveVisuoTactile} proposed an active pipeline that is able to reduce impact forces using a visual-tactile approach.

Robotic skins are generally a promising asset in the future of human-robot interaction. With an increase in the number of available whole-body robotic skins~\cite{cheng2019comprehensive}, new approaches can be invented to control robots or handle collisions. Similarly, \etal{Svarny} performed an extensive set of experiments to show the impact of robotic skin on collision forces~\cite{svarnyAirskin2022}.  \etal{Guadarrama-Olvera}~\cite{Olvera2019ComplianceSkin} proposed an approach for a pressure-driven controller of a humanoid robot based on the external force and also the size of the contact. An interesting subcategory are neuro-inspired robotic skins. An example of this category is the skin by \etal{Liu}~\cite{liu2022PrintedSynapticTransistor} which consists of synaptic transistors. The authors demonstrated that the skin is able to acquire pain-like reflexes. Moreover, human skin also has the ability to sense thermal stimuli. \etal{Neto}~\cite{neto2022SkinInspiredThermoreceptorsBasedElectronic} developed thermoreceptors-based skin that is able to learn at the hardware level to respond to temperature impulses. We refer the readers to the review~\cite{liu2022neuro} for more details on bio-inspired robotic sensing.

However, current robots are rarely equipped with human-inspired skins. Therefore, researchers are proposing human-like pain sensing for robots with more standard skins, or even without them. In~\cite{feng2022BraininspiredRobotPain}, the authors proposed \ac{snn} based on the Free Energy Principle to simulated pain in humanoid robot without any robotic skin. Similarly, also with spiking neurons, \etal{Kuehn}~\cite{kuehn2017ArtificialRobotNervous} presented \ac{arns} that consists of several layers of artificial neurons of different types. With this \ac{arns}, they were able to control a robotic arm. 

This work demonstrates how adaptive collision sensitivity thresholds can bring safety and productivity under one hood. 

%% file: Sections/method.tex
\section{Method}
We design and evaluate several ways of statically or dynamically setting the sensitivity or collision thresholds (``artificial robot pain'') in different parts of the robot. This Section describes the task performed by the robot in which we performed our tests, the possible reactions to the collision and the different settings for the threshold.

\subsection{Robot and Skin}
We use the UR10e 6-DOF robotic arm. The robot is a collaborative one, i.e., it can detect collisions and other distortions by itself. We selected the least restrictive setting of its safety system, so that all collisions are firstly detected by the sensitive skin and our program and skin and not the robot controller from the manufacturer. We control the real robot through its official \ac{ros} Noetic driver in velocity mode, i.e., we send velocities directly to individual joints of the robot. 

The robot is also equipped with a robotic skin  \airskin{}\footnote{\url{https://www.airskin.io/}}. \airskin{} detects changes in pressure inside each of its parts---we call them pads. Our robot with 11 pads can be seen in \figref{fig:main_fig}. By default, the system should be connected directly to the robot safety controller, and only binary information (collision / no collision) is available. With a special airskin{} maintenance controller, we can connect the skin to our computer and read the pressure values of the individual pads with the frequency of \SI{30}{\hertz}. The skin readout is also integrated to \ac{ros}. The standard operation of \airskin{}, which triggers robot stop after contact is detected through the dedicated safety lines, was not used here.

\subsection{Simulation, Software and Data}
Next to the real robot setup, we also created a simulation for both the robot and the skin. The simulation uses PyBullet~\cite{coumans2021} as a background physics engine. We created a front-end interface that contains the UR10e robot with simulated \airskin{}. It can work in two modes:
\begin{itemize}
    \item \textit{native mode}, where user has only low-level control of the robot without any external planners or controllers. The simulation can run off-screen or with a custom visualizer created with Open3D Python library~\cite{Zhou2018Open3d}. In this mode, the simulation speed is orders of magnitude faster than real-time (depending on the hardware) and can be used, for example, for reinforcement learning; 
    \item \textit{\ac{ros} mode}, where the simulation is integrated into \ac{ros} and thus provides the same controllers like the real robot and motion planner interfaces like MoveIt! can be easily used. Here, the simulation is artificially limited to run in real time, so it is as similar to the real-world setup as possible. Our experiments were performed in this mode. 
\end{itemize}

The skin is emulated using the distance of objects to virtual skin links on the robot. In the \ac{ros} mode of the simulation, an interface is available to hit specific skin pads with a specific effort. The simulation can be found at: \url{https://github.com/ctu-vras/adaptive-skin}.

\subsection{Performed Task}
\label{sec:task}
\begin{figure}
    \centering
    \includegraphics[width=0.75\columnwidth]{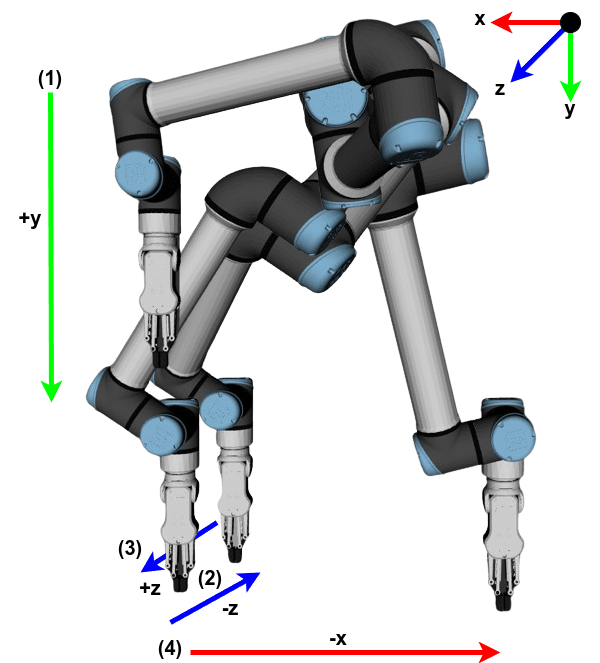}
    \caption{Top-view of the task performed by the robot. Order of movements is depicted by numbers in brackets.}
    \label{fig:task}
\end{figure}

We simulated a pick-and-place task, whereby the robot motion is disturbed by collisions, simulating physical human-robot collaboration. The robot first moves in its y-axis (forward from the base), then in the negative and positive z-axis (up and down), and finally in the negative x-axis (sideways)---see \figref{fig:task} and the video at \url{https://youtu.be/kWt2J1Qgzpk}. The task has been designed in this way, as it can be both theoretically and empirically shown that the collision forces will depend not only on the link speed but also on the collision direction and position in the workspace (see e.g., \cite{svarny20213d,svarnyAirskin2022}). All joints of the robot are employed during the task.

In every moment, we assume that every potential contact is a transient one. The ISO/TS 15066~\cite{ISOTS15066} described this as a \enquote{dynamic impact}, where the human body part impacted by the robot is not constrained and can retract from the robot, resulting in a shorter impact duration.

The robot is controlled in velocity mode, i.e., we send directly the velocity commands to the individual joints. This allows quick reactions to collisions---see \secref{sec:collision_reaction}.

\subsection{Collision Reaction}
\label{sec:collision_reaction}
The most straightforward reaction to a collision is to stop the robot. We also selected stop---later referred to as \stop{}---as one of our reaction policies. Velocity of \SI{0}{\radian\per\second} is sent to every joint when a collision is detected. The movement continues when the collision disappears.

Other possible reactions include going away from the contact, going around the contact point, or switching the robot to gravity compensation mode~\cite{Haddadin2008CollisionDetectionReaction}. We selected the option to move away from the contact point (\avoid{}). This policy is straightforward and, furthermore, the UR controller employs a similar strategy by default---the robot \enquote{bumps away} from the contact after a collision; see~\cite{svarnyAirskin2022}. The robot is controlled in velocity mode, thus we utilize the \ac{rrmc}~\cite{whitneyRRMC}. In other words, given 6D Cartesian velocity of a link (3 translation and 3 rotation velocities) $\dot{\mathbf{x}}$, the Jacobian matrix $\mathbf{J(q)}$, and the equation
\begin{equation}
    \dot{\mathbf{x}} = \mathbf{J(q)}\dot{\mathbf{q}},
\end{equation}
we can compute the desired joint velocities $\dot{\mathbf{q}}$ in every iteration of the algorithm as 
\begin{equation}
    \dot{\mathbf{q}} = \mathbf{J(q)}^{-1}\dot{\mathbf{x}}.
\end{equation}

After impact, the link of the robot that collided with the human needs to be retracted. When moving away from the collision, we want to keep the rotation of the given end-effector link the same. Thus, we set the rotational velocity to the vector $(0, 0, 0)^T$. The translation velocity is set to be a vector opposite to the direction vector of the link on which the skin was activated. 
The Jacobian matrix, $\mathbf{J(q)}$, will have a proper inverse only if it is square and full rank, i.e., in our case only for the end effector link away from singular configurations (6 joints and 6 dimensions of the Cartesian avoidance task). This cannot be guaranteed in general (all but the end effector link have fewer joints that can be recruited). Therefore,  we use the Moore-Penrose pseudoinverse instead.

\subsection{Collision Sensitivity Thresholds}
\label{sec:onset_of_pain}
We considered three collision sensitivity thresholds and four different ways of distributing them on the skin parts (pads) on the robot body: 
\begin{enumerate*}[label=(\arabic*)]
    \item the same threshold on every pad (\same{});
    \item different thresholds on pads (\diff{});
    \item adaptive threshold based on ISO/TS 15066 (\norm{});
    \item adaptive threshold based on ISO/TS 15066 and effective mass of the robot (\mass{}).
\end{enumerate*}

Settings (1) and (2) are static, i.e., the same threshold is set during the whole task. Settings (3) and (4) are dynamic and the thresholds for individual skin pads are set with the frequency of \SI{25}{\hertz} (every \SI{40}{\milli\second}). The choice of the frequency is affected by the time needed to compute the dynamic properties of the robot.

\subsubsection{\same{}}
In this setting,  all skin pads have the same threshold, specifically the most sensitive. We consider this metric as a baseline as it is a common setting when protective sensitive skins are employed.

\subsubsection{\diff{}}
We  divided the robot into three parts, similar to the human arm---upper arm, lower arm (forearm), and hand. This division is not only similar to humans, but also intuitive when looking at the robot, as it has two longer sections and then the gripper---see \figref{fig:main_fig}. We examined the velocities of the given body parts during the task and observed that the hand and lower arm usually move with a higher velocity than the upper arm. Therefore, we set the thresholds for this mode as the most sensitive for the hand and lower arm, and the middle threshold for the upper arm.

\subsubsection{\norm{}}
This settings of pad onsets is based on the equation for maximal velocity from ISO/TS 15066. The equation determines the maximal velocity using a mass-spring-mass model as
\begin{equation}
    v \leq  \frac{F_\mathrm{max}}{\sqrt{k}}\sqrt{m_{R}^{-1} + m_{H}^{-1}},
    \label{eq:v_pfl_orig}
\end{equation}
where $k$ is the spring constant of impacted human body, $m_R$ is the effective mass of the robot, $m_H$ is the effective mass of the impacted human body, $v$ is the Cartesian velocity of the robot link and $F_{max}$ is the maximal exerted force. The maximal force $F_{max}$ can be computed as
\begin{equation}
    \label{eq:f_norm}
    F_{max} = \frac{v\sqrt{k}}{\sqrt{m_{R}^{-1} + m_{H}^{-1}}}.
\end{equation}

We use \equationref{eq:f_norm} to compute the possible force exerted when colliding with each skin pad whenever a new state of the robot is available (and hence every link will have a new velocity). Again, based on ISO/TS 15066 we set $k=\SI{75000}{\newton\per\meter}$ (spring constant of the back of non-dominant hand; the most sensitive human body part according to \cite{ISOTS15066}, except for the head) and $m_R$ to the half of the moving mass of the robot up to the given skin part. Then, based on the maximal allowed force for transient contact (\SI{280}{N}) we selected the sensitivity thresholds $T$ for the given pad as:
\begin{equation}
\label{eq:force_to_thr}
T=
\begin{cases}
  0, & \text{if } F \geq \SI{280}{\newton} \\
  1, & \text{if } \SI{140}{\newton} \leq F < \SI{280}{\newton} \\
  2, & \text{if } F < \SI{140}{\newton},
\end{cases}
\end{equation}
where $0, 1, 2$ stands for the most, medium and less sensitive thresholds, respectively.

\subsubsection{\mass{}}
This property relates the current configuration and dynamics of the robot. Based on the work of Khatib~\cite{khatib1995inertial}, one can compute the effective mass $m_{\mathbf{u}}$ in the direction $\mathbf{u}$ as
\begin{equation}
    m_{\mathbf{u}} = \frac{1}{\mathbf{u}^T \mathbf{\Lambda}_v(\mathbf{q})^{-1} \mathbf{u}},
\end{equation}
where $\mathbf{\Lambda}_v(\mathbf{q})^{-1}$ is the upper $3\times 3$ matrix of Cartesian kinetic energy matrix $\mathbf{\Lambda}(\mathbf{q})^{-1}$, that can be computed as 
\begin{equation}
    \mathbf{\Lambda}(\mathbf{q})^{-1} = \mathbf{J(q)}\mathbf{M(q)}\mathbf{J(q)}^T,
\end{equation}
where $\mathbf{M(q)}$ is the joint space inertia matrix. 

In~\cite{ISOTS15066}, the $m_R$ used in \equationref{eq:f_norm} is estimated to be half the moving mass of the robot. This estimate is usually too conservative, and the actual effective mass is usually lower, resulting in less exerted force---see~\cite{kirschner2021effmass} for more discussion on the topic.

We use the computed $m_{\mathbf{u}}$ as the mass of the robot $m_R$ in \equationref{eq:f_norm}. The individual thresholds for skin pads are then, again, set using the computed force and \equationref{eq:force_to_thr}.

%% file: Sections/experiments.tex
\section{Experiments and Results}
We conducted the same set of experiments in two setups---in the real world and in simulation---with the four sensitivity threshold settings. We tested two scenarios---stopping (\stop{}) after contact and moving away from the contact (\avoid{}). We saved all data from the robot (including skin pressures) in the form of rosbags---available at \url{https://osf.io/3adrm/}.

\subsection{Experiment Description}
In both environments, we assume transient contact with a human body part with a weight of \SI{5.6}{\kilo\gram} (corrresponding to the human arm; weight $m_H$ in \equationref{eq:f_norm}) and we set the Cartesian velocity of the end-effector link to \SI{500}{\milli\meter\per\second}. This velocity was selected as the borderline velocity that in some configurations exceeds the \SI{280}{\newton} limit of ISO/TS 15066 when computed using \equationref{eq:f_norm}.

During the experiments, the robot always performs its task described in \secref{sec:task} and we let it collide with an obstacle. In each run, only one collision can happen. We examine collision during movement in each direction of the task (+y, -z, and -x axis) and on each of the body parts of the robot (upper arm, lower arm, and hand). As there are several skin pads on each of the body parts, we always select one pad that moves in the most similar direction compared to the end-effector of the robot. For each threshold setting (see \secref{sec:onset_of_pain}) there are 9 individual runs (1 per each axis; 3 per body part), making 36 runs in total per reaction behavior (\stop{} and \avoid{}). We performed 10 repetitions of each experiment in the simulation and 3 repetitions in the real world, resulting in 720 and 216 runs, respectively.

\subsection{Simulation Experiments}
\label{sec:sim_exps}
In this setup, the obstacle is a virtual ball that touches the skin. We selected the three thresholds for collision detection (see \secref{sec:onset_of_pain}) such that the most sensitive threshold can detect the collision all the time and then the other two thresholds correspond to 75\% and 95\% of the pressure range of the simulated skin.

The simulation setup is more controllable and allows for more precise comparison. All experiments were repeated 10 times. In \stop{}, the ball always appears exactly at the same time and stays in the environment for a maximal time of \SI{1}{\sec}. In other words, the maximum time that the robot can remain stopped is \SI{1}{\sec}, depending on the reaction time. For \avoid{}, the ball stays in the environment until the skin detects a collision, i.e., the distance traveled by the robot away from the collision point depends on the link Cartesian velocity and the collision sensitivity threshold.

\begin{figure}[htb]
    \centering
    \includegraphics[width=\columnwidth,trim={2cm 0 2cm 0},clip]{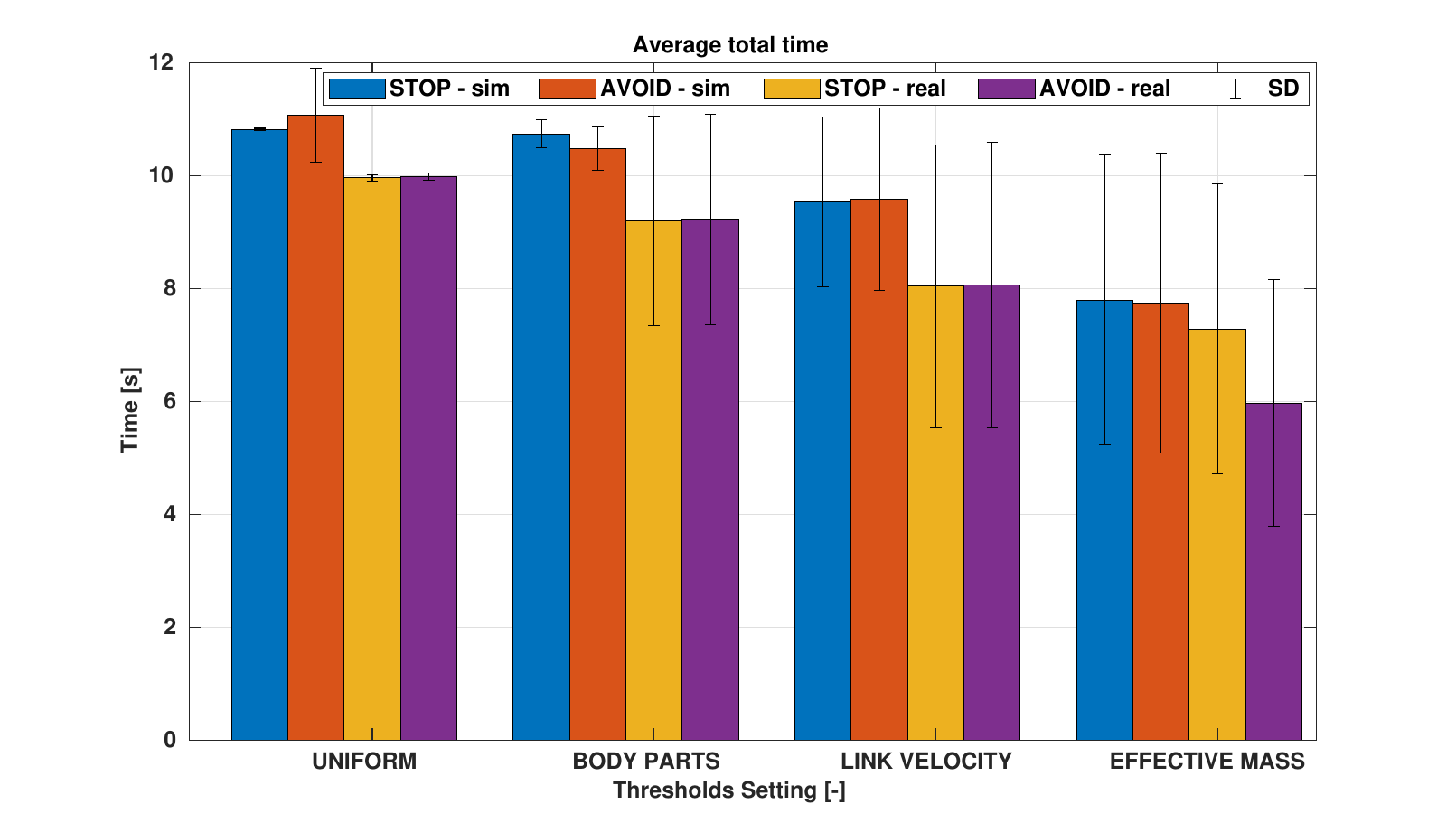}
    \caption{Average total run times $\pm$ 1 \acf{sd}. The values are averaged over all runs of the experiment in the given threshold setting.}
    \label{fig:total_time}
\end{figure}

\begin{table}[htb]
\centering
\begin{tabular}{cc|cc|}
\cline{3-4}
\multicolumn{2}{l|}{\multirow{2}{*}{}} & \multicolumn{2}{c|}{\textbf{Scenario}} \\ \cline{3-4} 
\multicolumn{2}{l|}{} & \multicolumn{1}{c|}{\textbf{Stop}} & \textbf{Avoid} \\ \hline
\multicolumn{1}{|c|}{\multirow{4}{*}{\textbf{sim}}} & \textbf{\same{}} & \multicolumn{1}{c|}{\SI{10.81}{\second}} & \SI{11.07}{\second} \\ \cline{2-4} 
\multicolumn{1}{|c|}{} & \textbf{\diff{}} & \multicolumn{1}{c|}{\SI{10.74}{\second}} & \SI{10.48}{\second} \\ \cline{2-4} 
\multicolumn{1}{|c|}{} & \textbf{\norm{}} & \multicolumn{1}{c|}{\SI{9.53}{\second}} & \SI{9.58}{\second} \\ \cline{2-4} 
\multicolumn{1}{|c|}{} & \textbf{\mass{}} & \multicolumn{1}{c|}{\SI{7.78}{\second}} & \SI{7.75}{\second} \\ \hline
\multicolumn{1}{|c|}{\multirow{4}{*}{\textbf{real}}} & \textbf{\same{}} & \multicolumn{1}{c|}{\SI{9.96}{\second}} & \SI{9.98}{\second} \\ \cline{2-4} 
\multicolumn{1}{|c|}{} & \textbf{\diff{}} & \multicolumn{1}{c|}{\SI{9.20}{\second}} & \SI{9.22}{\second} \\ \cline{2-4} 
\multicolumn{1}{|c|}{} & \textbf{\norm{}} & \multicolumn{1}{c|}{\SI{8.04}{\second}} & \SI{8.06}{\second} \\ \cline{2-4} 
\multicolumn{1}{|c|}{} & \textbf{\mass{}} & \multicolumn{1}{c|}{\SI{7.28}{\second}} & \SI{5.97}{\second} \\ \hline
\end{tabular}
\caption{Average total time. The values are averaged over all runs of the experiment in the given scenario. }
\label{tab:total_time}
\end{table}

The most important performance measure is the average run-time of the experiments based on the thresholds used. We provide bar graphs in \figref{fig:total_time} and numerical values in \tabref{tab:total_time}. In the simulated experiments, when a collision exceeding the sensitivity threshold was detected, the robot stopped for \SI{1}{\second}. However, in a real collaborative task, such events typically result in longer task interruptions, because an operator needs to re-enable the robot motion. Therefore, for performance evaluation, we add \SI{5}{\second} to each run when the robot detects a collision.

We can see that the highest time is for \same{} and the lowest for \mass{}. This outcome is something we expected and is closely related to \tabref{tab:stops_per_body_part}, where the percentage of cases where the robot reacts to the collision is shown. We will describe \stop{} and \avoid{} separately, as the different reaction policies influence the results.

\begin{table}[htb]
\centering
\begin{tabular}{cc|ccc|}
\cline{3-5}
\multicolumn{2}{l|}{\multirow{2}{*}{}} & \multicolumn{3}{c|}{\textbf{Body Part}} \\ \cline{3-5} 
\multicolumn{2}{l|}{} & \multicolumn{1}{c|}{\textbf{Upper}} & \multicolumn{1}{c|}{\textbf{Lower}} & \textbf{Hand} \\ \hline
\multicolumn{1}{|c|}{\multirow{4}{*}{\textbf{sim}}} & \textbf{\same{}} & \multicolumn{1}{c|}{100\%} & \multicolumn{1}{c|}{100\%} & 100\% \\ \cline{2-5} 
\multicolumn{1}{|c|}{} & \textbf{\diff{}} & \multicolumn{1}{c|}{100\%} & \multicolumn{1}{c|}{100\%} & 100\% \\ \cline{2-5} 
\multicolumn{1}{|c|}{} & \textbf{\norm{}} & \multicolumn{1}{c|}{71.67\%} & \multicolumn{1}{c|}{100\%} & 100\% \\ \cline{2-5} 
\multicolumn{1}{|c|}{} & \textbf{\mass{}} & \multicolumn{1}{c|}{46.6\%} & \multicolumn{1}{c|}{60.0\%} & 63.3\% \\ \hline
\multicolumn{1}{|c|}{\multirow{4}{*}{\textbf{real}}} & \textbf{\same{}} & \multicolumn{1}{c|}{100\%} & \multicolumn{1}{c|}{100\%} & 100\% \\ \cline{2-5} 
\multicolumn{1}{|c|}{} & \textbf{\diff{}} & \multicolumn{1}{c|}{56.6\%} & \multicolumn{1}{c|}{100\%} & 100\% \\ \cline{2-5} 
\multicolumn{1}{|c|}{} & \textbf{\norm{}} & \multicolumn{1}{c|}{27.8\%} & \multicolumn{1}{c|}{72.2\%} & 89.9\% \\ \cline{2-5} 
\multicolumn{1}{|c|}{} & \textbf{\mass{}} & \multicolumn{1}{c|}{27.8\%} & \multicolumn{1}{c|}{50.0\%} & 27.8\% \\ \hline
\end{tabular}
\caption{Percentage of cases when the robot reacted to collisions. Each value is averaged over \stop{} and \avoid{} scenarios.}
\label{tab:stops_per_body_part}
\end{table}

\begin{table}[htb]
\centering
\begin{tabular}{cc|ccc|}
\cline{3-5}
\multicolumn{2}{l|}{\multirow{2}{*}{}} & \multicolumn{3}{c|}{\textbf{Threshold}} \\ \cline{3-5} 
\multicolumn{2}{l|}{} & \multicolumn{1}{c|}{\textbf{0}} & \multicolumn{1}{c|}{\textbf{1}} & \textbf{2} \\ \hline
\multicolumn{1}{|c|}{\multirow{4}{*}{\textbf{sim}}} & \textbf{\same{}} & \multicolumn{1}{c|}{100 \%} & \multicolumn{1}{c|}{0 \%} & 0 \% \\ \cline{2-5} 
\multicolumn{1}{|c|}{} & \textbf{\diff{}} & \multicolumn{1}{c|}{66.6 \%} & \multicolumn{1}{c|}{33.3 \%} & 0 \% \\ \cline{2-5} 
\multicolumn{1}{|c|}{} & \textbf{\norm{}} & \multicolumn{1}{c|}{35.6 \%} & \multicolumn{1}{c|}{36.6 \%} & 27.8 \% \\ \cline{2-5} 
\multicolumn{1}{|c|}{} & \textbf{\mass{}} & \multicolumn{1}{c|}{0 \%} & \multicolumn{1}{c|}{37.7 \%} & 62.29 \% \\ \hline
\multicolumn{1}{|c|}{\multirow{4}{*}{\textbf{real}}} & \textbf{\same{}} & \multicolumn{1}{c|}{100 \%} & \multicolumn{1}{c|}{0 \%} & 0 \% \\ \cline{2-5} 
\multicolumn{1}{|c|}{} & \textbf{\diff{}} & \multicolumn{1}{c|}{66.6 \%} & \multicolumn{1}{c|}{33.3 \%} & 0 \% \\ \cline{2-5} 
\multicolumn{1}{|c|}{} & \textbf{\norm{}} & \multicolumn{1}{c|}{22.2 \%} & \multicolumn{1}{c|}{55.6 \%} & 22.2 \% \\ \cline{2-5} 
\multicolumn{1}{|c|}{} & \textbf{\mass{}} & \multicolumn{1}{c|}{0 \%} & \multicolumn{1}{c|}{35.2 \%} & 64.8 \% \\ \hline
\end{tabular}
\caption{Percentage of thresholds (0 to 2; most to least sensitive) in different modes in simulation and the real world. The numbers are averaged over \stop{} and \avoid{} scenarios.}
\label{tab:thr_percentage}
\end{table}

In case of \stop{}, the total time for \same{} and \diff{} is basically the same. This is supported by the fact that for these, the robot reacted every time, as can be seen in \tabref{tab:stops_per_body_part}. In the same table, we can find that for \mass{} and \norm{}, the robot stopped around 90\% and 57\%, respectively. It caused a difference between these two and the first two modes.

In case of \avoid{}, there is a small difference even between \same{} and \diff{} (about \SI{0.5}{\second}), which is probably caused by the fact that because the reaction time is higher, the direction of movement of the given link can be slightly different. It may not seem as a significant influence on total time, but because the only body part with different sensitivity threshold is the upper arm, the avoid movement is constrained only to the first two joints of the robot. Thus, even a small difference in movement direction can change avoidance velocity. The rest is analogous to the \stop{} scenario.

For both reaction strategies, the \acp{sd} are higher for dynamic threshold switching strategies. This would be expected, as these two have a higher deviation of stopped and uninterrupted runs.

\begin{figure}[htb]
    \centering
    \includegraphics[width=\columnwidth,trim={0.9cm 0 2cm 0},clip]{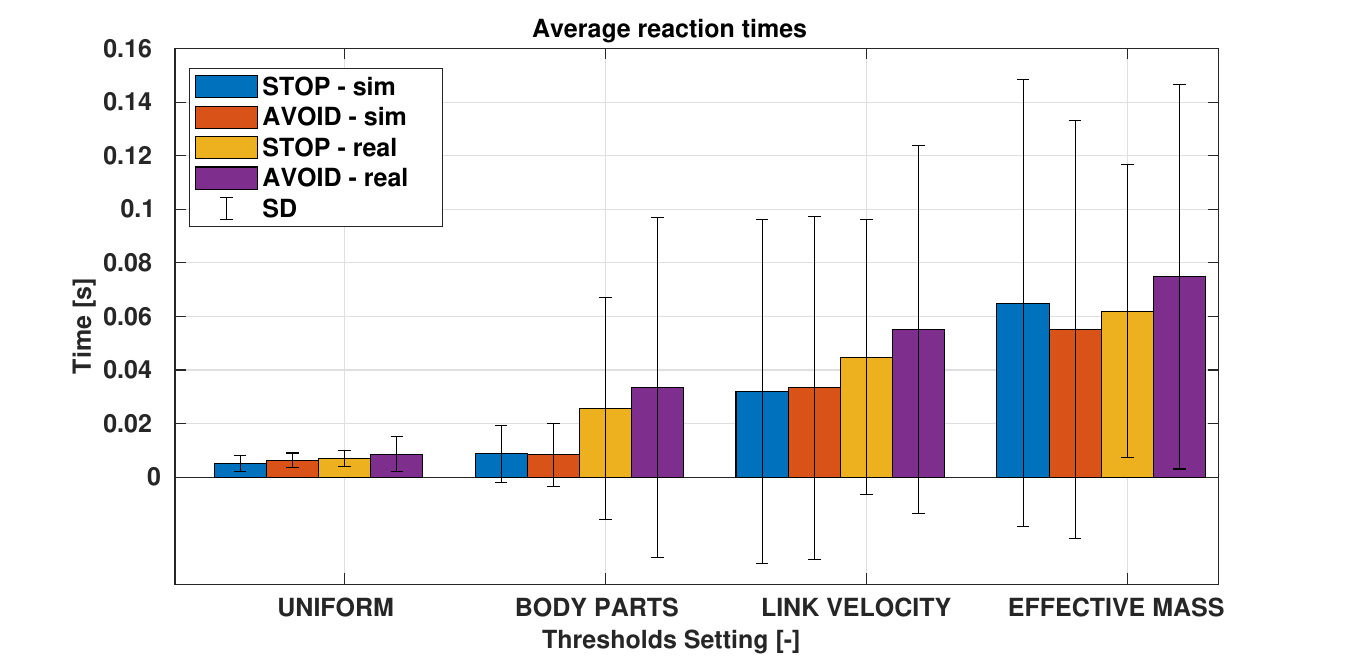}
    \caption{Average reaction times $\pm$ 1 \acf{sd}. The values are computed from runs of the experiments in which the robot actually reacted to collision.}
    \label{fig:reaction_times}
\end{figure}

Another property we can examine is the reaction time to collision. The values are shown in \figref{fig:reaction_times}. The results correspond to our hypothesis, that is, they should be the lowest for \same{} and the highest for \mass{}. The reason is that with \same{}, all thresholds are set to the most strict ones, thus the robot should stop faster. On the other hand, with \norm{} and \mass{} it is expected that there will be less strict thresholds in most cases, naturally resulting in higher reaction times. We can confirm this assumption using \tabref{tab:thr_percentage} or \figref{fig:main_fig}, where we can see that for \norm{} and \mass{}, threshold 0 is set only in 35\% and 0\% of cases, respectively. The overall longest reaction time is \SI{0.29}{\second} in \norm{} mode. It may seem high at first glance, but the correlation coefficient between reaction times and the force estimated using \equationref{eq:f_norm} is -0.86. It shows a high negative correlation, that is, that the estimated force is higher for the short reaction times and vice versa. I The average estimated force for cases where the reaction time is higher than \SI{0.2}{\second} is \SI{60}{\newton}, which is far below the limit of \SI{280}{\newton}. 

\begin{figure}[htb]
    \centering
    \includegraphics[width=\columnwidth,trim={0.9cm 0 2cm 0},clip]{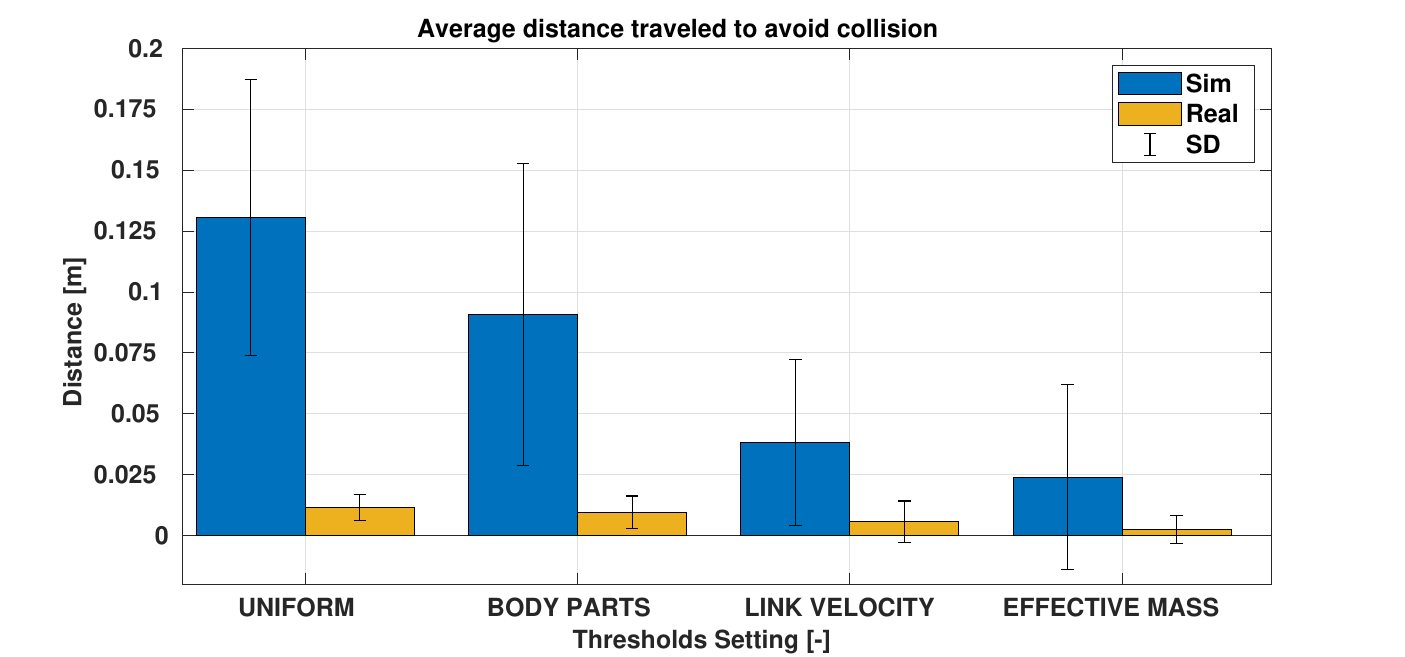}
    \caption{Average distance moved to avoid collision. The values are averaged over all runs of the experiment in the given scenario.}
    \label{fig:travel_dist}
\end{figure}

The last property we explore is the effect of thresholds on distance needed to go away from the collision---naturally, this is only evaluated in \avoid{}. The results are in \figref{fig:travel_dist}. We can see that, similarly to the total time, the traveled distance is the highest for \same{} and the lowest for \mass{}. Because \norm{} and \mass{} assign higher sensitivity thresholds (see \tabref{tab:thr_percentage} and \figref{fig:main_fig}), the robot needs to travel a shorter distance to get away from the collision. In our case, the robot immediately started to come back to trajectory and continue the task, and thus the hypothetically impacted human would have less time to move because of the shorter distance. However, we assume that in the real world there would be, again, some time when the robot would remain stopped and then a shorter avoidance distance means fewer interruptions of the task.

\subsection{Real Setup}
\begin{figure}[htb]
    \centering
    \begin{subfigure}[t]{0.49\columnwidth}
        \centering
        \includegraphics[width=1\textwidth,]{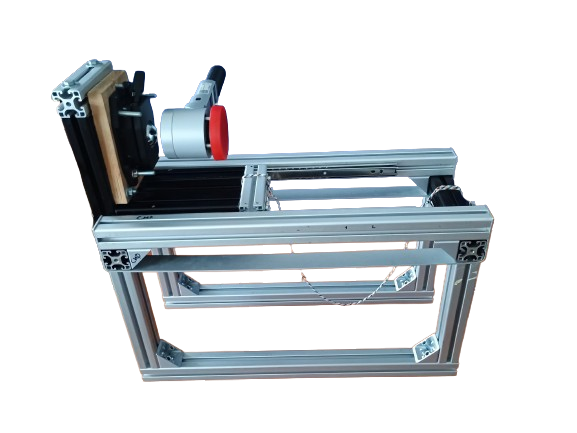}
        \caption{Transient contact simulation construction.}
        \label{fig:transient_setup}
    \end{subfigure}
    \begin{subfigure}[t]{0.41\columnwidth}
        \centering
        \includegraphics[width=1\textwidth]{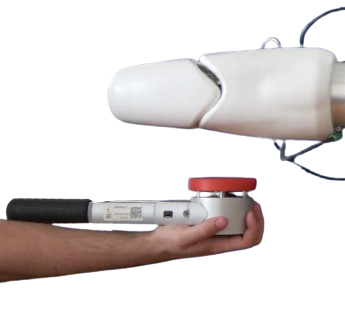}
        \caption{CBSF-75-Basic force measuring device in the hand of the participant.}
        \label{fig:force_meter}
    \end{subfigure}
    \caption{The \textit{CBSF-75-Basic} measuring device mounted on the transient contact setup and in the hand of a participant.}
    \label{fig:force_measurements}
    \customVspace{-1em}
\end{figure}

In the real world, the impacts were recorded with a human experimenter. To find AIRSKIN pressure values approximately corresponding to the force thresholds, we recorded several values with \textit{CBSF-75-Basic}, which is a certified measuring device for validation of collaborative tasks with the range \SIrange{20}{500}{\newton}. To simulate transient contact, we furthermore used a construction consisting of moving mass and bearing sliders---\figref{fig:transient_setup}. The mass of the whole moving mass (used for \equationref{eq:f_norm}) is \SI{5.6}{\kilo\gram}. We kindly refer the reader to~\cite{svarnyAirskin2022} for more details on the construction. Using this setup, we measured impact forces in x- and y-axis movements of our experiments in hand and lower body regions of the robot. We cannot measure forces when moving in the negative z-axis, as gravity would pull the moving mass to the ground, and we were unable to measure forces on upper arm because of limited space not allowing to properly hit the measuring device. Based on the measurement, we selected the most sensitive threshold as a value that always reacted to the collision with the device; the second threshold as the last value that reacted and the least sensitive threshold circa as 90\% of the pressure range of the skin. 

All other experiments were conducted with a real human. We performed 3 repetitions for every impact. The human being hit by the robot was instructed to let the robot hit his hand, while being as \enquote{inattentive} to the robot as possible. The aim is to simulate real collisions in the factory. The human also carried the \textit{CBSF-75-Basic} measuring device in his hand, basically as an \enquote{extension} of his arm---see \figref{fig:force_meter}. Based on~\cite{ISOTS15066}, the weight of human hand and lower arm is about \SI{2.6}{\kilo\gram}. Together with the measuring device, the weight can be estimated to be the same as that of the transient measuring setup. Furthermore, we used the measuring device to see whether individual touches were comparable as real-world contacts are not as controllable as in the simulation. The average deviation between individual repetitions was \SI{10}{\newton}. Given the certified measuring error of the device (\SI{3}{N}), we find this error to be reasonably low. However, the authors still acknowledge that measuring the impacts in this way is not as rigorous as it could be with some more sophisticated setup.

The purpose of the real-world experiments here is mainly to assess the results obtained in the simulation. If we again start with total time, we can see that the times are in general lower than in the simulation. It is caused by the fact that the collision in the real world lasts only a short amount of time (we show some examples in the accompanying video) compared to the fixed amount of \SI{1}{\second} in the simulation (see \secref{sec:sim_exps} for details). Unlike in simulation, there is even a difference between \same{} and \mass{}. It is supported by \tabref{tab:stops_per_body_part} where it is visible that, compared to the simulation, the robot is now stopped even in the \diff{} mode. Moreover, the difference between \stop{} and \avoid{} for \mass{} is higher than in the simulation. The reason lies in the lower number of repetitions (3 versus 10 in the simulation), which results in a higher weight of each repetition on the result. Also, it could be caused by the fact that the real experiments were not as controlled as in the simulation---see discussion in the previous paragraph. However, in general, the trends are the same as in the simulation.

In case of reaction times, \figref{fig:reaction_times} also shows trends similar to those for the simulation. The reaction times are higher than in the simulation, which is probably caused by a difference in the deformation of the real skin. Furthermore, looking at \tabref{tab:stops_per_body_part}, we can see that the robot reacted less often in the real world (70.1\% vs. 86.8\%). We attribute this mainly to the fact that the sensitivity of individual pads differs slightly among the whole body. And because we estimated pressure thresholds based on impact on the hand and lower arm, the sensitivity of the upper arm was worse, as visible in the table. However, this is a skin-based issue and is beyond the scope of this work.

The last property is the traveled distance to avoid the collision shown in \figref{fig:travel_dist}. Again, the trend is similar to the simulation with the highest distance for \same{} and the lowest for \mass{}. Compared to simulation, the distances are generally about ten times lower. The reason is similar to a small difference in total time, that is, the impact duration in the real world is much smaller. In the future, we would advise, for example, to set a minimum avoidance time to avoid this problem. 

Overall, we can see that the real-world results show the same trends as the simulation and prove the results are consistent over the different environments.


%% file: Sections/conclusion.tex
\section{Conclusion, Discussion, and Future Work}

We conducted an extensive set of experiments in simulation, which we created, and in the real world, to find out how collision sensitivity thresholds robots should be set in order to warrant safety and maximize productivity. The experiments showed that the sensitivity thresholds determine the proportion of contacts that do not interrupt robot task execution and also collision detection reaction time. These two indirectly influence task properties like total task time or distance needed to travel away from collision. In the beginning, we asked two questions: \textit{Should there be different sensitivity (``onsets of pain'') thresholds for different parts of the robot skin? Based on what criterion should the onsets be set?} The answer to the first one is straightforward---yes, the collision sensitivity thresholds should be different. Individual links of the robot move with different velocity and have different weight, therefore they have capability to exert a different force during impact. The visualization of changes in thresholds over time during the task can be seen in \figref{fig:main_fig}. Using different thresholds helps mainly to keep the task uninterrupted and smooth in case of small, harmless collisions.

The answer to the second question is more complicated. Based on our experiment, we can recommend the following. If the task is simple (for example, the robot moves only in one axis), record and examine the maximal velocities (and exerted forces) of individual links of the robot and set the thresholds based on them. In case of more complex tasks, we recommend using adaptive thresholds. We tested two, one based on the maximum force estimation using the equation from ISO/TS 15066 (\norm{}) and one based on the same equation but replacing the fixed robot weight in the equation with the computed robot weight that is based on the current configuration (\mass{}). We prefer the second option, as the default equation from ISO/TS 15066 tends to overestimate the force and thus lower the pain thresholds. This is nicely visible in \figref{fig:main_fig}, where we can that \mass{} sets less sensitive thresholds in more cases. 

\begin{table}[htb]
\centering
\begin{tabular}{l|l|l|l|l|}
\cline{2-5}
 & Hand x & Hand y & Lower x & Lower y \\ \hline
\multicolumn{1}{|l|}{\norm{}} & \SI{277.6}{\newton} & \SI{276.4}{\newton} & \SI{275.8}{\newton} & \SI{292.9}{\newton} \\ \hline
\multicolumn{1}{|l|}{\mass{}} & \SI{82.2}{\newton} & \SI{242.3}{\newton} & \SI{224.6}{\newton} & \SI{209.2}{\newton} \\ \hline
\multicolumn{1}{|l|}{Real} & \SI{74.2}{\newton} & \SI{125.2}{\newton} & \SI{76.8}{\newton} & \SI{80.8}{\newton} \\ \hline
\end{tabular}
\caption{Real and estimated force for hand and lower arm during movement x- and y-axis. The values are averaged over 5 repetitions.}
\label{tab:real_vs_estimated}
\end{table}

We also measured forces in the real world to further prove the results. In \tabref{tab:real_vs_estimated}, we provide estimated forces for both modes and real measured force when the robot moves in x- and y-axis measured on hand and lower arm parts of the robot. We can see that real forces are much lower than both estimates, but \mass{} is slightly more precise. 
Accurate estimation of the effective mass depends on the accuracy of the model of the robot dynamics and other factors and may be difficult to obtain in practice. Empirical, \textit{in situ}, measurements are need for safety to be guaranteed (see \cite{svarny20213d,svarnyAirskin2022}).

The work presented here is the first step toward setting artificial skin sensitivity thresholds adaptively. This article was mainly motivated by collaborative robot applications where safety needs to be guaranteed by limiting the collision forces but at the same time maximum robot speed is desired. We demonstrated the potential of different ``pain threshold'' settings on the 6-axis manipulator (UR10e) with AIRSKIN. There are several directions of future work. First, the approaches we devised here can be extended to even more skin parts like when covering a complete body of a humanoid robot. Sensitivity thresholds for additional body parts can be derived from \cite{ISOTS15066}. Second, the ``pain thresholds'' and the corresponding avoidance reactions could be more closely inspired by humans (e.g., the withdrawal reflex) or even learnable (e.g., \cite{liu2022neuro}). Third, the sensitivity of the body parts to touch could be designed for social human interaction. Heslin~\cite{heslin1983} studied the pleasantness vs. intrusiveness of touch between sexes in relation to a stranger, a friend, and a close friend in the United States. The body areas where touch was rated as pleasant or unpleasant strongly depend on the combination of these factors. More context-awareness on the part of the robot will be needed. 


%% file: main.bbl
\begin{thebibliography}{10}
\providecommand{\url}[1]{#1}
\csname url@rmstyle\endcsname
\providecommand{\newblock}{\relax}
\providecommand{\bibinfo}[2]{#2}
\providecommand\BIBentrySTDinterwordspacing{\spaceskip=0pt\relax}
\providecommand\BIBentryALTinterwordstretchfactor{4}
\providecommand\BIBentryALTinterwordspacing{\spaceskip=\fontdimen2\font plus
\BIBentryALTinterwordstretchfactor\fontdimen3\font minus \fontdimen4\font\relax}
\providecommand\BIBforeignlanguage[2]{{%
\expandafter\ifx\csname l@#1\endcsname\relax
\typeout{** WARNING: IEEEtran.bst: No hyphenation pattern has been}%
\typeout{** loaded for the language `#1'. Using the pattern for}%
\typeout{** the default language instead.}%
\else
\language=\csname l@#1\endcsname
\fi
#2}}

\bibitem{mittendorfer2011humanoid}
P.~Mittendorfer and G.~Cheng, ``{Humanoid Multimodal Tactile-Sensing Modules},'' \emph{IEEE Transactions on Robotics}, vol.~27, no.~3, pp. 401--410, 2011.

\bibitem{mittendorfer2012integrating}
------, ``{Integrating Discrete Force Cells into Multi-modal Artificial Skin},'' in \emph{2012 12th IEEE-RAS International Conference on Humanoid Robots (Humanoids 2012)}.\hskip 1em plus 0.5em minus 0.4em\relax IEEE, 2012, pp. 847--852.

\bibitem{park2024_LowCostSkin}
K.~Park, K.~Shin, S.~Yamsani, K.~Gim, and J.~Kim, ``Low-cost and easy-to-build soft robotic skin for safe and contact-rich human–robot collaboration,'' \emph{IEEE Transactions on Robotics}, vol.~40, pp. 2327--2338, 2024.

\bibitem{cannata2008embedded}
G.~Cannata, M.~Maggiali, G.~Metta, and G.~Sandini, ``{An Embedded Artificial Skin for Humanoid Robots },'' in \emph{2008 IEEE International Conference on Multisensor Fusion and Integration for Intelligent Systems}.\hskip 1em plus 0.5em minus 0.4em\relax IEEE, 2008, pp. 434--438.

\bibitem{maiolino2013flexible}
P.~Maiolino, M.~Maggiali, G.~Cannata, G.~Metta, and L.~Natale, ``{A Flexible and Robust Large Scale Capacitive Tactile System for Robots},'' \emph{IEEE Sensors Journal}, vol.~13, no.~10, pp. 3910--3917, 2013.

\bibitem{Goncalves2022_Punyo}
A.~Goncalves, N.~Kuppuswamy, A.~Beaulieu, A.~Uttamchandani, K.~M. Tsui, and A.~Alspach, ``Punyo-1: Soft tactile-sensing upper-body robot for large object manipulation and physical human interaction,'' in \emph{2022 IEEE 5th International Conference on Soft Robotics (RoboSoft)}, 2022, pp. 844--851.

\bibitem{cheng2019comprehensive}
G.~Cheng, E.~Dean-Leon, F.~Bergner, J.~Rogelio Guadarrama~Olvera, Q.~Leboutet, and P.~Mittendorfer, ``{A Comprehensive Realization of Robot Skin: Sensors, Sensing, Control, and Applications},'' \emph{Proceedings of the IEEE}, vol. 107, no.~10, pp. 2034--2051, 2019.

\bibitem{kandel1991principles}
E.~Kandel, J.~Schwartz, and T.~Jessell, \emph{{Principles of Neural Science}}.\hskip 1em plus 0.5em minus 0.4em\relax Appleton \& Lange, 1991.

\bibitem{ISOTS15066}
``{ISO/TS 15066 Robots and robotic devices -- Collaborative robots},'' International Organization for Standardization, Geneva, CH, Tech. Rep., 2016.

\bibitem{onsetOfPain}
``{Research project No. FP-0317: Collaborative robots – Investigation of pain sensibility at the Man-Machine-Interface},'' Institute for Occupational, Social and Environmental Medicine at the Johannes Gutenberg University, Mainz, Germany, Tech. Rep., 2014.

\bibitem{Haddadin2016}
S.~Haddadin and E.~Croft, ``{Physical Human-Robot Interaction},'' in \emph{Springer Handbook of Robotics}.\hskip 1em plus 0.5em minus 0.4em\relax Springer, 2016, pp. 1835--1874.

\bibitem{Villaini2018Survey}
V.~Villani, F.~Pini, F.~Leali, and C.~Secchi, ``{Survey on human–robot collaboration in industrial settings: Safety, intuitive interfaces and applications},'' \emph{Mechatronics}, vol.~55, pp. 248--266, 2018.

\bibitem{Fischer2023CollisionTests}
C.~Fischer, M.~Neuhold, M.~Steiner, T.~Haspl, M.~Rathmair, and S.~Schlund, ``{Collision Tests in Human-Robot Collaboration: Experiments on the Influence of Additional Impact Parameters on Safety},'' \emph{IEEE Access}, vol.~11, pp. 118\,395--118\,413, 2023.

\bibitem{kirschner2021effmass}
R.~J. Kirschner, N.~Mansfeld, G.~G. Peña, S.~Abdolshah, and S.~Haddadin, ``{Notion on the Correct Use of the Robot Effective Mass in the Safety Context and Comments on ISO/TS 15066},'' in \emph{2021 IEEE International Conference on Intelligence and Safety for Robotics (ISR)}, 2021, pp. 6--9.

\bibitem{lucci2020combining}
N.~Lucci, B.~Lacevic, A.~M. Zanchettin, and P.~Rocco, ``{Combining Speed and Separation Monitoring With Power and Force Limiting for Safe Collaborative Robotics Applications},'' \emph{IEEE Robotics and Automation Letters}, vol.~5, no.~4, pp. 6121--6128, 2020.

\bibitem{svarny2022functional}
P.~Svarny, M.~Hamad, A.~Kurdas, M.~Hoffmann, S.~Abdolshah, and S.~Haddadin, ``{Functional Mode Switching for Safe and Efficient Human-Robot Interaction},'' in \emph{2022 IEEE-RAS 21st International Conference on Humanoid Robots (Humanoids)}, 2022, pp. 888--894.

\bibitem{Xu2023ActiveVisuoTactile}
C.~Xu, Y.~Zhou, B.~He, Z.~Wang, C.~Zhang, H.~Sang, and H.~Liu, ``{An Active Strategy for Safe Human–Robot Interaction Based on Visual–Tactile Perception},'' \emph{IEEE Systems Journal}, vol.~17, no.~4, pp. 5555--5566, 2023.

\bibitem{svarnyAirskin2022}
P.~Svarny, J.~Rozlivek, L.~Rustler, M.~Sramek, {\"O}.~Deli, M.~Zillich, and M.~Hoffmann, ``{Effect of Active and Passive Protective Soft Skins on Collision Forces in Human--Robot Collaboration},'' \emph{Robotics and Computer-Integrated Manufacturing}, vol.~78, p. 102363, Dec. 2022.

\bibitem{Olvera2019ComplianceSkin}
J.~R. Guadarrama-Olvera, E.~Dean-Leon, F.~Bergner, and G.~Cheng, ``{Pressure-Driven Body Compliance Using Robot Skin},'' \emph{IEEE Robotics and Automation Letters}, vol.~4, no.~4, pp. 4418--4423, 2019.

\bibitem{liu2022PrintedSynapticTransistor}
F.~Liu, S.~Deswal, A.~Christou, M.~Shojaei~Baghini, R.~Chirila, D.~Shakthivel, M.~Chakraborty, and R.~Dahiya, ``{Printed Synaptic Transistor--Based Electronic Skin for Robots to Feel and Learn},'' \emph{Science Robotics}, vol.~7, no.~67, p. eabl7286, June 2022.

\bibitem{neto2022SkinInspiredThermoreceptorsBasedElectronic}
J.~Neto, R.~Chirila, A.~S. Dahiya, A.~Christou, D.~Shakthivel, and R.~Dahiya, ``Skin-{{Inspired Thermoreceptors-Based Electronic Skin}} for {{Biomimicking Thermal Pain Reflexes}},'' \emph{Advanced Science}, vol.~9, no.~27, p. 2201525, 2022.

\bibitem{liu2022neuro}
F.~Liu, S.~Deswal, A.~Christou, Y.~Sandamirskaya, M.~Kaboli, and R.~Dahiya, ``Neuro-inspired electronic skin for robots,'' \emph{Science robotics}, vol.~7, no.~67, p. eabl7344, 2022.

\bibitem{feng2022BraininspiredRobotPain}
H.~Feng and Y.~Zeng, ``{A Brain-Inspired Robot Pain Model Based on a Spiking Neural Network},'' \emph{Frontiers in Neurorobotics}, vol.~16, Dec. 2022.

\bibitem{kuehn2017ArtificialRobotNervous}
J.~Kuehn and S.~Haddadin, ``An {{Artificial Robot Nervous System To Teach Robots How To Feel Pain And Reflexively React To Potentially Damaging Contacts}},'' \emph{IEEE Robotics and Automation Letters}, vol.~2, no.~1, pp. 72--79, Jan. 2017.

\bibitem{coumans2021}
E.~Coumans and Y.~Bai, ``Pybullet, a python module for physics simulation for games, robotics and machine learning,'' \url{http://pybullet.org}, 2016--2021.

\bibitem{Zhou2018Open3d}
Q.-Y. Zhou, J.~Park, and V.~Koltun, ``{{Open3D}: {A} Modern Library for {3D} Data Processing},'' \emph{arXiv:1801.09847}, 2018.

\bibitem{svarny20213d}
P.~Svarny, J.~Rozlivek, L.~Rustler, and M.~Hoffmann, ``{3D Collision-Force-Map for Safe Human-Robot Collaboration},'' in \emph{2021 IEEE International Conference on Robotics and Automation (ICRA)}, 2021, pp. 3829--3835.

\bibitem{Haddadin2008CollisionDetectionReaction}
S.~Haddadin, A.~Albu-Schaffer, A.~De~Luca, and G.~Hirzinger, ``{Collision Detection and Reaction: A Contribution to Safe Physical Human-Robot Interaction},'' in \emph{2008 IEEE/RSJ International Conference on Intelligent Robots and Systems}, 2008, pp. 3356--3363.

\bibitem{whitneyRRMC}
D.~E. Whitney, ``{Resolved Motion Rate Control of Manipulators and Human Prostheses},'' \emph{IEEE Transactions on Man-Machine Systems}, vol.~10, no.~2, pp. 47--53, 1969.

\bibitem{khatib1995inertial}
O.~Khatib, ``{Inertial Properties in Robotic Manipulation: An Object-Level Framework},'' \emph{The International Journal of Robotics Research}, vol.~14, no.~1, pp. 19--36, 1995.

\bibitem{heslin1983}
R.~Heslin, T.~D. Nguyen, and M.~L. Nguyen, ``{Meaning of touch: The case of touch from a stranger or same sex person},'' \emph{Journal of Nonverbal Behavior}, vol.~7, pp. 147--157, 1983.

\end{thebibliography}
